\documentclass[lettersize,journal]{IEEEtran}
\usepackage{amsmath,amsfonts}
\usepackage{algorithmic}
\usepackage{algorithm}
\usepackage{array}
\usepackage[caption=false,font=normalsize,labelfont=sf,textfont=sf]{subfig}
\usepackage{textcomp}
\usepackage{stfloats}
\usepackage{url}
\usepackage{verbatim}
\usepackage{graphicx}
\usepackage{cite}
\begin{document}
\title{A Master-to-Simulation Interface with Demonstration-Trajectory Guidance for Vascular Intervention}
\author{Mingyu Zou, Shuxiang Guo,~\IEEEmembership{Fellow,~IEEE,}, Sheng Cao, Pengcheng Li, Jian Guo, Bin Wang, and Mingchao Ding
        % <-this % stops a space
\thanks{Research supported by the Shenzhen "Medical Engineering Integration" Special Support Program Project under Grant F-2024-Z99-502365, Shenzhen, China; in part by the High level of special funds from Southern University of Science and Technology under Grant G03034K003, Shenzhen, China. (Corresponding author: Shuxiang Guo.)}% <-this % stops a space
\thanks{S. Guo is with the Department of Electronic and Electrical Engineering, Southern University of Science and Technology, Shenzhen, Guangdong 518055, China, and also with the School of Life Science and the Key Laboratory of Convergence Medical Engineering System and Healthcare Technology, Ministry of Industry and Information Technology, Beijing Institute of Technology, Beijing 100081, China. (e-mail: guo.shuxiang@sustech.edu.cn).}
\thanks{M. Zou and S. Cao are with the Beijing Institute of Technology, Beijing 100081, China.}
\thanks{P. Li is with the Department of Electronic and Electrical Engineering, Southern University of Science and Technology, Shenzhen, Guangdong 518055, China.}
\thanks{J. Guo is with Shenzhen Institute of Advanced Biomedical Robot Company Limited, Shenzhen 518000, Guangdong Province, China.}
\thanks{B. Wang and M. Ding are with Aerospace Center Hospital and the Key Laboratory of Convergence Medical Engineering System and Healthcare Technology, Beijing Institute of Technology, Beijing 100081, China.}}

% The paper headers
%\markboth{Journal of \LaTeX\ Class Files,~Vol.~14, No.~8, August~2021}%
%{Shell \MakeLowercase{\textit{et al.}}: A Sample Article Using IEEEtran.cls for IEEE Journals}
%\IEEEpubid{0000--0000/00\$00.00~\copyright~2021 IEEE}
% Remember, if you use this you must call \IEEEpubidadjcol in the second
% column for its text to clear the IEEEpubid mark.
\maketitle
\begin{abstract}
Robotic vascular intervention can reduce radiation exposure to physicians and improve manipulation stability, but effective guidance remains challenging because conventional fluoroscopic images provide limited spatial information. This paper developed a vascular intervention interactive system that connects a real robotic master controller with the SOFA(Simulation Open Framework Architecture) physical simulation platform and reused successful operation trajectories as visual guidance in physical phantom experiments. Experiments were conducted to validate the master-to-SOFA control loop and to compare three guidance conditions: no guidance, centerline guidance, and successful-trajectory guidance. The control validation results showed reliable command delivery under velocity-mode operation. In the physical phantom experiment, centerline guidance showed no obvious improvement over no guidance. In contrast, successful trajectory guidance provided a clear performance benefit. Compared with the no-guidance condition, it reduced task completion time, collision number, rollback number, and rotation number by approximately \(66.8\%\), \(34.3\%\), \(55.6\%\), and \(66.7\%\), respectively. These results indicate that successful trajectories obtained under instrument-vessel interaction can provide practical guidance cues, offering a simple and effective visual assistance strategy for robotic vascular intervention experiments.
\end{abstract}
\begin{IEEEkeywords}
Vascular intervention robot; master-follower control; trajectory record; visual guidance.
\end{IEEEkeywords}
\section{Introduction}
\IEEEPARstart{I}{ntravascular} intervention has become an effective treatment for cardiovascular and cerebrovascular diseases, owing to its advantages of minimal trauma, reduced postoperative complications, and rapid recovery\cite{palaniappan20262026, yao2025advancing, mandhumula2025robotic, cao2024elastic}. During such procedures, catheters and guidewires are advanced through narrow and tortuous vessels to reach target lesions under fluoroscopic or digital subtraction angiography (DSA) guidance. However, conventional manual intervention highly depends on the operator's experience, tactile sensitivity, and hand-eye coordination. Maintaining stable and precise manipulation during long procedures remains challenging. In addition, interventional physicians are exposed to ionizing radiation for prolonged periods, which poses non-negligible occupational health risks\cite{fu2023magnetically, cao2023reciprocating}. To address these limitations, vascular interventional robotic systems have been developed. By allowing operators to manipulate instruments from a remote console behind radiation shielding, robotic systems can reduce radiation exposure, improve motion stability, and enhance manipulation precision\cite{ning2023robotic}.

Although robotic systems can improve the stability and repeatability of catheter and guidewire manipulation, physicians still need to interpret fluoroscopic or digital subtraction angiography images and continuously determine the spatial relationship between the instruments and the surrounding vasculature. Reinok et al\cite{reinok2024automated}. proposed an automated passive tracking method for MR-guided endovascular interventions that detects discrete passive markers on an MR-visible guidewire in real time. The detection results are integrated into a custom software interface, offering a cost-effective, radiation-free, and near-real-time guidance assistance method for MR-guided interventions. Zhang et al\cite{zhang2025hdcar}. proposed HDCAR, a 3D-2D registration network for abdominal aortic vessels that aligns preoperative CTA-based 3D vascular models with intraoperative DSA images. This method effectively compensates for missing vascular structures in single-frame DSA images, providing more precise imaging guidance and assistance for endovascular aneurysm repair procedures. However, these methods do not provide a structured opportunity for less experienced operators to explore and practice surgical measures in advance. The experience gained from successful virtual surgeries has not been transferred to subsequent actual surgeries. This limitation is important for young or less experienced physicians. Before performing an actual surgery, they may need to understand not only the anatomical route, but also how the guidewire and catheter should be advanced, rotated, retracted, and coordinated when passing through difficult vascular regions.

\begin{figure*}[!h]
	\centerline{\includegraphics[width=2\columnwidth]{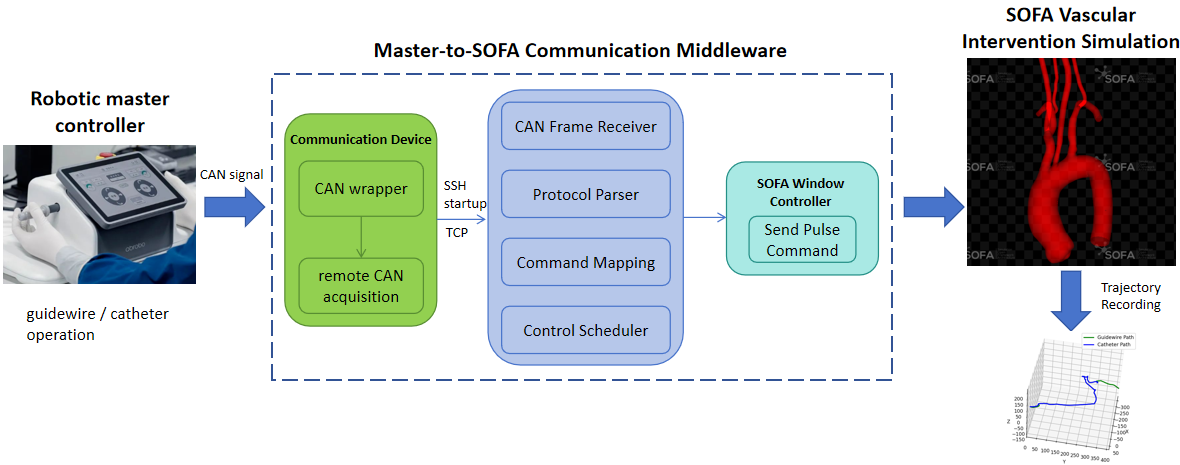}}
	\caption{Overview of the interactive intervention system.}
	\label{fig1}
\end{figure*}

A successful surgical trajectory is more than a geometric curve. It implicitly records the physical response of the instrument during guidewire-vessel interaction, including bending, contact-induced deviation, passive compliance, and feasible advancement direction under operational constraints. If such successful trajectories can be collected and reused, they may provide operators with prior knowledge that is unavailable from vessel geometry alone. This successful trajectory containing prior knowledge can serve as a practical guidance reference, reflecting not only where the guidewire should go, but also how a successful route progresses through the vascular structure under physical constraints. 

This study aims to enable doctors to conduct virtual training before actual surgery and transform successful operational experience into interpretable visual references. The main contributions are summarized as follows. First, a vascular intervention interactive system is developed by connecting a real robotic master controller with the SOFA physical simulation platform, enabling real-time guidewire/catheter manipulation and complete trajectory recording. Second, a successful trajectory acquisition and reuse method is proposed, in which successful trajectories containing instrument-vessel interaction information are used as visual guidance, mapping the recorded 3D successful trajectory onto real-time images of a physical vascular phantom. Finally, comparative experiments are conducted under the same vascular model, including no-guidance operation, centerline-guidance operation, and successful-trajectory-guidance operation, to evaluate the effectiveness of the proposed guidance strategy.

\section{SYSTEM DESIGN AND IMPLEMENTATION}
This study implemented an interactive intervention system based on a robotic master manipulator and SOFA. In Fig. 1, this system interfaces the physical vascular interventional robotic master manipulator with the SOFA physical simulation platform. Through the master manipulator, the operator can actuate the virtual guidewire and catheter within the simulation environment in real time. Concurrently, the system comprehensively records the 3D trajectories of the guidewire and catheter tips under physical contact and deformation constraints. After obtaining successful trajectories, a physical vascular phantom experiment is introduced to evaluate whether the successful trajectory can serve as an effective visual guidance cue\cite{li2025human, LI2025108731}.

\subsection{Robotic Master manipulator}
\begin{figure}[!h]
	\centerline{\includegraphics[width=1\columnwidth]{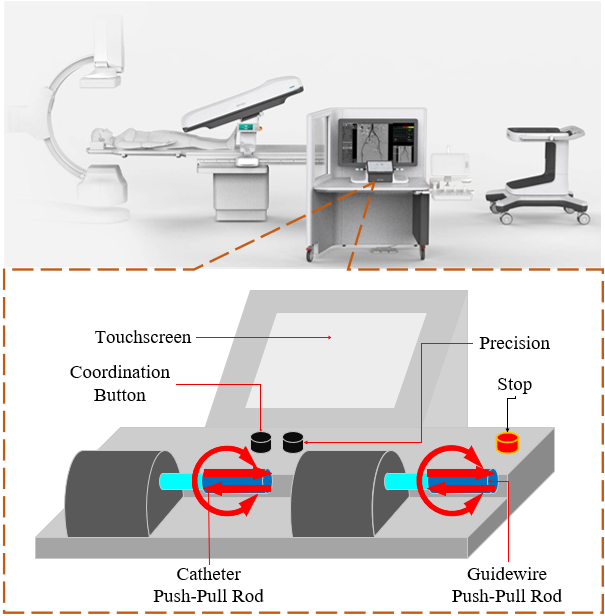}}
	\caption{Master manipulator of vascular intervention robot.}
	\label{fig2}
\end{figure}
The vascular interventional robot used in this study is illustrated in Fig.2. The system is designed to support remote manipulation of interventional instruments while separating the operator from the radiation field. During operation, the physician controls the master manipulator outside the X-ray exposure area, and the follower robotic unit drives the catheter and guidewire at the operating side. This configuration reduces occupational radiation exposure and can also lessen the physical burden associated with wearing heavy lead protection during long procedures.
The platform consists mainly of a master controller and a follower delivery robot. The master side is designed according to the manipulation habits of interventional physicians. Instead of using a conventional joystick or knob-based interface, it adopts a dual push-pull lever structure to reproduce the advancing, retracting, and cooperative manipulation of catheters and guidewires. The follower robot is responsible for executing the commands from the master manipulator and delivering multiple interventional instruments in a coordinated manner\cite{yan2025novel, lyu2024deep, jin2023novel, yan2024novel}.

The master manipulator was used as the human-machine input device for the proposed guidance framework. The operator performed guidewire and catheter manipulation on the real master device, while the corresponding commands were transmitted to the virtual vascular intervention scene constructed in SOFA. In this way, successful guidewire movements generated under realistic master-side operation could be recorded in the simulation environment.

The master manipulator communicates through the Controller Area Network (CAN) protocol. To connect the physical master manipulator with the SOFA-based simulation program, CAN messages were converted into TCP packets through a CAN-to-Ethernet gateway. The converted data were then transmitted via a standard Ethernet interface. This communication structure allowed the physical robotic controller and the simulation environment to exchange control commands and instrument states in real time.
\subsection{SOFA-based Vascular Intervention Simulation}
The virtual intervention environment was developed using SOFA with the BeamAdapter plugin. SOFA provides a modular physical simulation framework suitable for interactive medical simulation, while BeamAdapter enables slender flexible instruments to be modeled using beam theory. In the proposed system, the guidewire and catheter were represented as deformable beam structures, allowing bending, twisting, and contact response during vascular intervention procedures to be simulated.

The vascular model was imported into SOFA as a rigid triangular mesh and used as the anatomical constraint for the intervention task. Collision response between the instruments and the vessel wall was enabled through triangle, line, and point collision models. The guidewire and catheter were modeled using Cosserat-beam-based components. Their mechanical behavior was solved using an implicit integration scheme, and the InterventionalRadiologyController was used to map external motion commands into instrument advancement, retraction, and rotation. Through this setup, the simulated guidewire could respond to vessel geometry and contact constraints rather than following an idealized geometric trajectory.

During each simulation step, the three-dimensional positions of the guidewire and catheter tips were recorded together with their command events. The trajectories of the two instruments were stored separately in CSV files, while command events were recorded with their timestamps, instrument identities, and action labels. These synchronized records enabled each trajectory sample to be associated with its corresponding instrument action. After a complete successful trial is obtained, the recorded historical data are processed to generate the reference trajectory used in subsequent visualization.

\subsection{Master-to-SOFA Communication Middleware}
In this study, a communication middleware was developed to connect the real robotic master controller with the SOFA-based vascular intervention simulation platform. The middleware acts as a bidirectional bridge between physical manipulation signals and virtual instrument motion. Its main functions include master-side signal acquisition, CAN frame parsing, command semantic mapping, SOFA window control, and real-time image display.

The overall workflow of the middleware is shown in Fig.2. First, the physical master controller generates CAN messages according to the operator's push-pull and rotational manipulation. The CAN data are collected from the master-side device through a remote CAN wrapper. In the implementation, an SSH connection is established to the master device, and a ZMQ-based communication client subscribes to the CAN message stream. ZMQ is a lightweight asynchronous messaging library that enables efficient communication between distributed processes. The raw CAN frames are then transmitted to the local computer through TCP-based ZMQ communication.
Let a received CAN frame at time \(t\) be denoted as
\begin{equation}
F_t = \{id_t, dlc_t, data_t\},
\end{equation}
where \(id_t\) is the CAN identifier, \(dlc_t\) is the data length, and \(data_t\) contains the encoded control payload. The protocol parser extracts the control address and two motion-related variables from the payload:
\begin{equation}
(a_t, d_{r,t}, d_{p,t}) = \Phi(data_t),
\end{equation}
where \(a_t\) denotes the control address, \(d_{r,t}\) denotes the rotational input, and \(d_{p,t}\) denotes the axial push-pull input. According to the address field, the middleware determines whether the current command belongs to the guidewire channel or catheter channel:
\begin{equation}
m_t =
\begin{cases}
guidewire, & a_t = a_{gw},\\
catheter, & a_t = a_{ct}.
\end{cases}
\end{equation}
The original master-side input is then converted into discrete manipulation semantics. Axial displacement is mapped to forward or backward motion, while rotational displacement is mapped to clockwise or counterclockwise rotation:
\begin{equation}
u_t =
\begin{cases}
forward, & d_{p,t} > 0,\\
backward, & d_{p,t} < 0,\\
rotate\_cw, & d_{r,t} > 0,\\
rotate\_ccw, & d_{r,t} < 0.
\end{cases}
\end{equation}
To avoid command jitter near the neutral position, a dead-zone mapping is applied to the input magnitude. The command intensity is computed as
\begin{equation}
I_t =
\begin{cases}
300, & |d_t| < 300,\\
|d_t|, & |d_t| \geq 300,
\end{cases}
\end{equation}
where \(d_t\) is selected from either \(d_{p,t}\) or \(d_{r,t}\) according to the active motion direction. The resulting command can be represented as
\begin{equation}
C_t = (m_t, u_t, I_t).
\end{equation}
The command intensity is then converted into a directional pulse command for SOFA control.

The simplified procedure of the middleware is described in Algorithm~\ref{alg:middleware}.
\begin{algorithm}[!htbp]
\small
\caption{Master-to-SOFA communication middleware}
\label{alg:middleware}
\begin{algorithmic}[1]
\STATE Initialize SOFA window controller
\STATE Establish SSH connection to the master device
\STATE Start ZMQ subscriber for CAN message acquisition
\STATE Register CAN callback function
\WHILE{system is running}
    \STATE Receive CAN frame \(F_t = \{id_t, dlc_t, data_t\}\)
    \STATE Parse payload: \((a_t, d_{r,t}, d_{p,t}) \leftarrow \Phi(data_t)\)
\IF{\(a_t = a_{gw}\) or \(a_t = a_{ct}\)}
        \STATE Determine controlled instrument \(m_t\)
        \STATE Determine motion direction \(u_t\)
        \STATE Compute command intensity \(I_t\)
        \STATE Generate command \(C_t = (m_t, u_t, I_t)\)
        \STATE Send command pulse to SOFA window
    \ENDIF
\ENDWHILE
\end{algorithmic}
\end{algorithm}

\subsection{Reference Trajectory Generation}
The recorded data were converted into a reference trajectory after a successful simulation trial was completed. The procedure was applied independently to the guidewire and catheter. The input for each instrument was a time-ordered sequence of three-dimensional tip samples and synchronized command labels.
For instrument $q\in\{gw,ct\}$, the raw record was represented as
\begin{equation}\mathcal{T}^{q}=\{(t_i,\mathbf{p}^{q}_i,a^{q}_i)\}_{i=1}^{N_q},\label{eq:raw_reference_trajectory}\end{equation}
where $t_i$ is the recording timestamp, $\mathbf{p}^{q}_i=(x^{q}_i,y^{q}_i,z^{q}_i)$ is the three-dimensional tip position, and $a^{q}_i$ is the action label associated with the sample. The action label was obtained from the synchronized command record and was used only to identify the manipulation state of each trajectory segment.
Invalid samples and repeated or near-repeated points are first removed. A point is retained when its displacement from the preceding retained point satisfies
\begin{equation}
\|\mathbf{p}_i-\mathbf{p}_{i-1}\|_2>\epsilon,
\end{equation}
where \(\epsilon\) is the minimum displacement threshold used to remove redundant samples.
The remaining trajectory is then smoothed using a moving-average filter to reduce small local fluctuations introduced by discrete simulation sampling. For an interior point, the smoothed position is calculated as
\begin{equation}
\hat{\mathbf{p}}_i=\frac{1}{2k+1}\sum_{j=-k}^{k}\mathbf{p}_{i+j},
\end{equation}
where \(k\) is the half-width of the smoothing window. At the two boundaries, the nearest valid samples are used to preserve the original start and end positions. Then, the smoothed points are resampled according to cumulative arc length to obtain a uniformly distributed trajectory.

Finally, the processed samples were sorted according to their action labels. These labels were used to distinguish the guidewire advancement segment and the catheter advancement, retraction, and rotation segments by color. Consecutive samples belonging to the same instrument and carrying the same action label were regarded as one action-specific trajectory segment. A new segment was created when the action label changed. The spatial coordinates of each segment were directly inherited from the corresponding recorded trajectory and displayed in different colors on the trajectory. The resulting color-coded trajectory can help the operator identify the expected instrument motion and the corresponding manipulation phase.

\section{Experiments and Results}
\subsection{Experimental Setup}
The experiments were conducted to evaluate the feasibility of the proposed system and the effectiveness of the successful-trajectory-based guidance strategy. All experiments were performed using the same vascular model, the same SOFA simulation scene, and the same robotic master controller. The operator manipulated the physical master controller to drive the virtual guidewire and catheter in SOFA. The trajectory of the guidewire and catheter tip were automatically recorded during each trial. A trial was considered completed when the guidewire tip reached the predefined target region.

The experiments consisted of two parts. First, a master-to-SOFA control validation experiment was conducted to verify whether the physical master controller could stably drive the SOFA simulation. Four basic manipulation commands were tested, including guidewire translation, guidewire rotation, catheter translation, and catheter rotation. Each command was repeated multiple times. In addition, a combined manipulation task was performed to evaluate continuous operation and instrument switching.
Second, for camera-based physical phantom validation, the same vascular geometry was used to construct a transparent vascular phantom. Comparative guidance experiments were designed under three conditions: no-guidance operation, centerline-guidance operation, and successful-trajectory-guidance operation. In Fig.3, all three conditions were carried out on the same physical phantom. In the centerline-guidance and successful-trajectory-guidance conditions, the corresponding overlays were manually aligned to the camera image and displayed as visual references, while the no-guidance condition used no additional overlay. The same target task and the same operator were used to ensure comparability.

The previously recorded command labels were used to divide each instrument trajectory into sequential action-specific trajectory segments. In the projected overlay, green denoted guidewire advancement, while blue, cyan, and pink denoted catheter advancement, retraction, and rotation, respectively. This color coded overlay serves as a visual assistance during navigation, providing additional action related information to the operator.

\begin{figure}[!h]
	\centerline{\includegraphics[width=1\columnwidth]{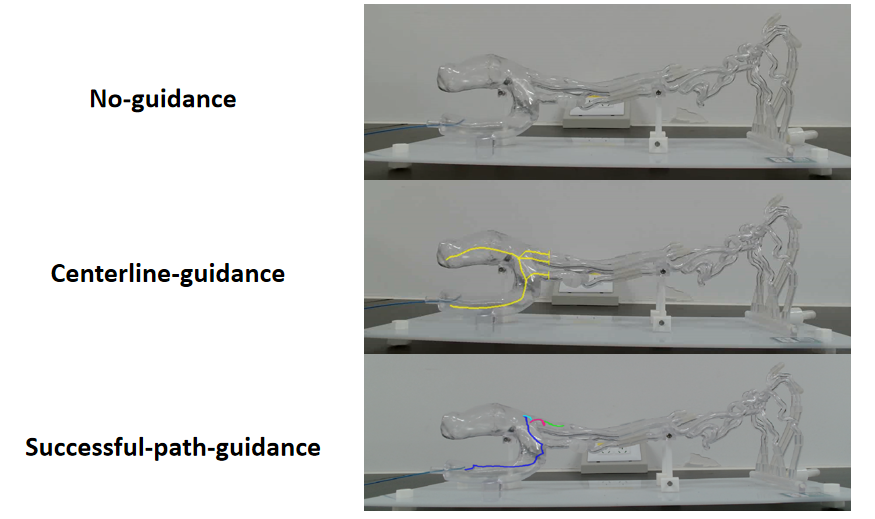}}
	\caption{Comparison of physical phantom guidance under three guidance conditions.}
	\label{fig3}
\end{figure}

The virtual intervention environment was implemented using SOFA v24.12 with the BeamAdapter plugin. The same aortic-arch geometry as the physical phantom was imported as a rigid triangular surface mesh and scaled by a factor of 4 in the simulation scene. Triangle, line, and point collision models were enabled for the vessel. The guidewire and catheter were modeled as deformable beam instruments. The radius of the guidewire is 0.45 mm and the Young's modulus is 6740. The radius of the catheter is 1.0 mm and the Young's modulus is 1000. Both instruments used a mass density of 5.0$\times10^{-8}$ in the simulation model.
The instrument mechanics were solved using an Euler implicit solver with Rayleigh stiffness and mass damping coefficients of 0.1. Contact response was handled using an LCP constraint solver with a friction coefficient of 0.1, a contact distance of 2, and a maximum of 20,000 iterations. The instrument collision proximity was set to 0.6. The interventional controller used a 5-unit command step and an initial exposed length of 50 for both instruments. The simulation GUI was configured at 1080$\times$1080 pixels, and the middleware exchanged instrument-state data with the SOFA scene at 20 Hz.

Both the SOFA scene and the physical phantom represented the same aortic-arch vascular anatomy. The transparent physical phantom was fixed on a rigid platform, and an external camera was placed at a fixed viewpoint to acquire the guidance image. The phantom, camera, insertion region, and target region were kept unchanged throughout the three guidance conditions. For the centerline-guidance and successful-trajectory-guidance conditions, the corresponding reference overlay was manually aligned with the camera image before the trials and then kept fixed during the repeated experiments. The no-guidance condition used the same phantom and camera configuration without an additional overlay.

\subsection{Evaluation Metrics}
\subsubsection{Metrics for Master-to-SOFA Interactive Control Validation}
For the master-to-SOFA interactive control validation, the command-to-pulse success rate and command latency were used to evaluate the stability and responsiveness of the communication loop. The command-to-pulse success rate was defined as the proportion of master-side commands that were correctly converted into valid SOFA control pulses without command loss or false triggering.

Command latency was defined as the elapsed time between the reception of a master-side signal and the generation of the corresponding SOFA control pulse. The mean, standard deviation, and maximum latency were reported to describe both the average response and the occasional latency peaks during operation.

\subsubsection{Metrics for Physical Phantom guidance Evaluation}
For the physical phantom guidance experiment, four metrics were used to evaluate the effectiveness of visual guidance: task completion time, collision number, rollback number, and rotation number. These metrics were selected according to the characteristics of the aortic arch traversal task.

The task completion time \(T_c\) was defined as the elapsed time from task start to successful completion. A shorter \(T_c\) indicates higher guidance efficiency.

The collision number \(N_{col}\) was defined as the number of visible contacts between the guidewire/catheter tip and the vessel wall that caused obvious motion obstruction or required subsequent adjustment.

The rollback number \(N_{back}\) was defined as the number of backward adjustment events during the guidance process. A rollback event indicates that the operator needed to withdraw the instrument to correct an unfavorable advancement state.

The rotation number \(N_{rot}\) was defined as the total number of rotational adjustment events required to complete the task. A smaller \(N_{rot}\) indicates that fewer angular corrections were required to pass through the curved vascular segment.

\subsection{Master-to-SOFA Interactive Control Validation}
To evaluate the reliability and real-time performance of the proposed master-to-SOFA communication layer, five validation experiments were conducted under velocity-mode operation, including guidewire translation, guidewire rotation, catheter translation, catheter rotation, and combined catheter-guidewire manipulation. Each single-axis task was repeated 20 times. In the combined-operation task, sequential guidewire advancement, rotation, withdrawal, catheter advancement, rotation, and withdrawal were performed to evaluate the system response under multi-instrument operation. For each trial, the command execution result in SOFA was observed and recorded. A command was regarded as successful if it was converted into a valid SOFA control pulse without command loss or false triggering.

The evaluation focused on the command transmission from the real robotic master manipulator to the SOFA simulation environment. For single-axis experiments, only the target-axis commands were included in the statistical analysis, while incidental cross-axis commands caused by operator-induced hand motion were excluded.

In Table I, all five validation experiments achieved a command-to-pulse success rate of 100.00\%, indicating that the proposed communication middleware could reliably transmit control commands from the physical master manipulator to the SOFA simulation scene. The mean latency of the five tasks ranged from 14.50 ms to 27.10 ms. The combined-operation task involved 1106 valid commands and achieved a mean latency of 21.64 ms, demonstrating that the system maintained stable command transmission during continuous multi-instrument manipulation.
Overall, the validation results demonstrate that the developed master-to-SOFA interactive control interface can support reliable velocity-mode manipulation. The system is capable of transmitting guidewire and catheter translation/rotation commands in real time and can remain operational during sequential combined manipulation.

\begin{table*}[htbp]
\centering
\caption{Master-to-SOFA control validation results under velocity-mode operation}
\begin{tabular}{llccccc}
\hline
Mode & Task & $N_{cmd}$ & $R_c$ (\%) & $\bar{\tau}$ (ms) & Std. of $\tau_i$ (ms) & Max. $\tau_i$ (ms) \\
\hline
Velocity & Guidewire translation & 300 & 100.00\% & 20.25 & 32.40 & 136.70 \\
Velocity & Guidewire rotation & 288 & 100.00\% & 15.28 & 31.96 & 173.39 \\
Velocity & Catheter translation & 259 & 100.00\% & 27.10 & 38.17 & 155.54 \\
Velocity & Catheter rotation & 270 & 100.00\% & 14.50 & 30.36 & 156.75 \\
Velocity & Combined operation & 1106 & 100.00\% & 21.64 & 36.94 & 211.55 \\
\hline
\multicolumn{7}{l}{\footnotesize $N_{cmd}$: number of valid commands; $R_c$: command-to-pulse success rate.} \\
\multicolumn{7}{l}{\footnotesize $\bar{\tau}$: mean command latency; $\tau_i$: latency of the $i$-th command.} \\
\end{tabular}
\end{table*}

\begin{table*}[htbp]
\centering
\caption{Physical phantom guidance results under different guidance conditions}
\begin{tabular}{lcccc}
\hline
Condition & \(T_c\) (s) & \(N_{col}\) & \(N_{back}\) & \(N_{rot}\) \\
\hline
No guidance & $20.8 \pm 4.7$ & $7.0 \pm 1.5$ & $3.6 \pm 1.3$ & $3.9 \pm 1.1$ \\
Centerline guidance & $18.4 \pm 4.1$ & $7.1 \pm 1.0$ & $3.3 \pm 0.7$ & $3.8 \pm 0.9$ \\
Successful trajectory guidance & $6.9 \pm 1.4$ & $4.6 \pm 0.5$ & $1.6 \pm 0.5$ & $1.3 \pm 0.5$ \\
\hline
\end{tabular}
\end{table*}

\subsection{Physical Phantom guidance Evaluation}
Five participants took part in the physical phantom guidance experiment. Each participant completed two trials under each of the three guidance conditions: no guidance, centerline guidance, and successful trajectory guidance. Therefore, a total of 30 trials were conducted. All participants performed the same aortic-arch guidance task using the same physical phantom, camera viewpoint, insertion position, and predefined target region. Before each trial, the guidewire and catheter were returned to the same initial configuration, and the corresponding visual guidance condition was presented to the participant. The trial ended when the guidewire tip reached the predefined target region. Task completion time, collision number, rollback number, and rotation number were recorded for each trial.

In Table II, The centerline-guidance condition showed no obvious improvement over the no-guidance condition, indicating that a purely geometric reference trajectory was insufficient for this curved-vessel task. In contrast, successful trajectory guidance substantially improved the guidance performance. Compared with no guidance, the mean task completion time decreased from \(20.8 \pm 4.7\) s to \(6.9 \pm 1.4\) s, corresponding to a reduction of approximately \(66.8\%\). The mean collision, rollback, and rotation numbers decreased from \(7.0 \pm 1.5\), \(3.6 \pm 1.3\), and \(3.9 \pm 1.1\) to \(4.6 \pm 0.5\), \(1.6 \pm 0.5\), and \(1.3 \pm 0.5\), representing reductions of approximately \(34.3\%\), \(55.6\%\), and \(66.7\%\), respectively.
Overall, these results indicate that centerline guidance did not lead to a substantial improvement in the aortic arch traversal task. Since the centerline does not reflect the actual guidewire motion tendency under contact constraints, it may be insufficient for guiding curved-vessel guidance. By contrast, the successful trajectory guidance provided a more task-specific visual cue derived from successful instrument motion, helping doctors make better decisions while improving efficiency, reducing contact, correcting frequency, and rotational adjustment.
\section{Conclusion}
This paper developed a vascular intervention interactive system that connects a real robotic master controller with the SOFA physical simulation platform and reused successful operation trajectories as visual guidance for physical phantom experiments. Through the developed master-to-SOFA communication middleware, guidewire and catheter translation and rotation commands were transmitted to the simulation environment in real time, enabling interactive manipulation and trajectory recording under vessel-contact constraints. The control validation experiments demonstrated reliable command delivery under velocity-mode operation, confirming that the proposed middleware can provide a stable basis for simulation-based trajectory acquisition. Furthermore, comparative experiments on a physical vascular phantom showed that centerline guidance did not produce an obvious improvement over no guidance in the aortic arch traversal task. In contrast, successful trajectory guidance reduced task completion time, collision number, rollback number, and rotation number by approximately \(66.8\%\), \(34.3\%\), \(55.6\%\), and \(66.7\%\), respectively. These results suggest that a successful trajectory obtained under instrument-vessel interaction can provide practical guidance cues for curved-vessel manipulation.

\bibliographystyle{IEEEtran}
\bibliography{IEEEabrv,refs}
\end{document}